%% file: main.tex
\documentclass[10pt,twocolumn,letterpaper]{article}
\usepackage{wrapfig}
\usepackage{amsmath}
\usepackage{stmaryrd} 
\newcommand{\btheta}{\boldsymbol{\theta}}
\newcommand{\bbeta}{\boldsymbol{\beta}}
\newcommand{\btau}{\boldsymbol{\tau}}
\newcommand{\bphi}{\boldsymbol{\phi}}
\newcommand{\bPhi}{\boldsymbol{\Phi}}

\newcommand{\bj}{\mathbf{j}}
\newcommand{\bJ}{\mathbf{J}}

\newcommand{\bV}{\mathbf{V}}
\newcommand{\bP}{\mathbf{P}}
\newcommand{\bv}{\mathbf{v}}
\newcommand{\ba}{\mathbf{a}}

\newcommand{\bc}{\mathbf{c}}

\newcommand{\ie}{\textit{i.e., }}
\newcommand{\eg}{\textit{e.g., }}
\newcommand{\etal}{\textit{et al. }}

\newcommand{\mcol}{m_{\text{col}}}
\newcommand{\mdist}{m_{\text{dist}}}
\newcommand{\mtouch}{m_{\text{touch}}}
\newcommand{\ours}{\textit{Ours}}
\newcommand{\vae}{\textit{VAE}}
\newcommand{\vaegan}{\textit{VAEGAN}}
\newcommand{\method}{\textit{MACS}}

\newcommand\blfootnote[1]{%
  \begingroup
  \renewcommand\thefootnote{}\footnote{#1}%
  \addtocounter{footnote}{-1}%
  \endgroup
}

\newcommand{\GT}{GT}
 \usepackage{cvpr}              
\usepackage[pagenumbers]{} 
\input{preamble}

\definecolor{cvprblue}{rgb}{0.21,0.49,0.74}
\usepackage[pagebackref,breaklinks,colorlinks,citecolor=cvprblue]{hyperref}

\def\paperID{323} 
\def\confName{3DV\xspace}
\def\confYear{2024\xspace}

\title{MACS: Mass Conditioned 3D Hand and Object Motion Synthesis}

\author{
Soshi Shimada$^{1,2,*}$  $\;\;\;\;$  
Franziska Mueller$^{3}$   $\;\;\;\;$  
Jan Bednarik$^{3}$   $\;\;\;\;$  
Bardia Doosti$^{3}$   $\;\;\;\;$  
Bernd Bickel$^{3}$    \\
Danhang Tang$^{3}$   $\;$  
Vladislav Golyanik$^{1}$   $\;$  
Jonathan Taylor$^{3}$   $\;$  
Christian Theobalt$^{1,2}$   $\;$  
Thabo Beeler$^{3}$  \\\\ 
$^{1}$MPI for Informatics, SIC$\;\;$
$^{2}$ VIA Research Center $\;\;$ 
$^{3}$ Google  
}

\begin{document}
\twocolumn[{
\vspace{-1cm}
\maketitle
\vspace{-0.8cm}
\begin{center}
  \centering
  \includegraphics[width=\textwidth]{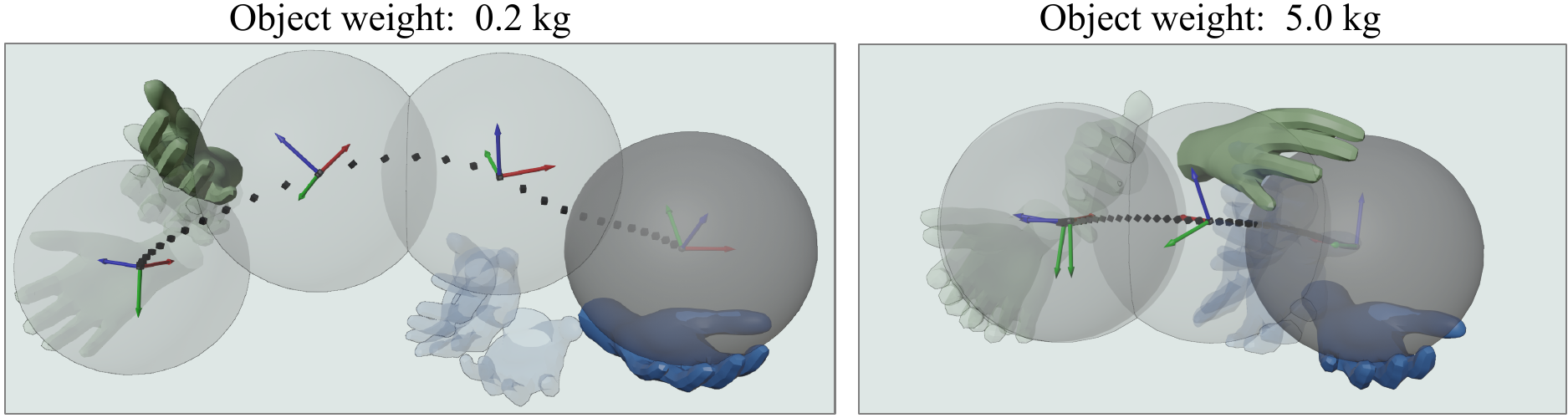} %
  \captionof{figure}{Example visualizations of 3D object manipulation synthesized by our method \method. Conditioning object mass values of $0.2$kg (left) and $5.0$kg (right) are given to the model for the action type "passing from one hand to another". \method\ plausibly reflects the mass value in the synthesized 3D motions.}\label{fig:teaser}
\end{center}
}]

\blfootnote{*Work done while at Google.}
\input{sec/0_abstract}     
\input{sec/1_intro}
\input{sec/2_related_works}
\input{sec/3_method}
\input{sec/4_experiments}
\input{sec/5_conclusions}
{
    \small
    \bibliographystyle{ieeenat_fullname}
    \bibliography{main}
}
\input{sec/6_appendix}
\end{document}

%% file: preamble.tex
\usepackage[dvipsnames]{xcolor}

%% file: sec/0_abstract.tex
 
\begin{abstract}
The physical properties of an object, such as mass, significantly affect how we manipulate it with our hands. Surprisingly, this aspect has so far been neglected in prior work on 3D motion synthesis. 
To improve the naturalness of the synthesized 3D hand-object motions, this work proposes \method\---the first \textit{MAss Conditioned 3D hand and object motion Synthesis} approach. 
Our approach is based on 
cascaded diffusion models and generates interactions that plausibly adjust based on the object's mass and interaction type. \method\ also accepts a manually drawn 3D object trajectory as input and synthesizes the natural 3D hand motions conditioned by the object's mass. 
This flexibility enables \method\ to be used for various downstream applications, such as generating synthetic training data for ML tasks, fast animation of hands for graphics workflows, and generating character interactions for computer games. 
We show experimentally that a small-scale dataset is sufficient 
for  \method\ to reasonably generalize across interpolated and extrapolated object masses unseen during the training. 
Furthermore, \method\ shows moderate generalization to unseen objects, thanks to the mass-conditioned contact labels generated by our surface contact synthesis model ConNet. Our comprehensive user study confirms that the synthesized 3D hand-object interactions are highly plausible and realistic. Project page link: \url{https://vcai.mpi-inf.mpg.de/projects/MACS/} 
\end{abstract}

%% file: sec/1_intro.tex
\section{Introduction}
\label{sec:intro} 
Hand-object interaction plays an important role in our daily lives, involving the use of our hands in a variety of ways such as grasping, lifting, and throwing. It is crucial for graphics applications (\eg AR/VR, avatar communication and character animation) to synthesize or capture physically plausible interactions for their enhanced realism. Therefore, there has been a growing interest in this field of research, and a significant amount of work has been proposed in grasp synthesis \cite{krug2010efficient, GRAB:2020,grady2021contactopt,graspingField2020,li2007data}, object manipulation \cite{ye2012synthesis,mordatch2012contact,christen2022d,ghosh2022imos,zhang2021manipnet}, 3D reconstruction \cite{wang2020rgb2hands,schroder2017hand,mueller2019real,corona2020ganhand,tekin2019h+,liu2021semi,Hu2022}, graph refinement \cite{pollard2005physically,detry2010refining,zhou2022toch} and contact prediction \cite{brahmbhatt2019contactdb}.%

Because of the high-dimensionality of the hand models and inconsistent object shape and topology, synthesizing plausible 3D hand-object interaction is challenging. Furthermore, errors of even a few millimeters can cause collisions or floating-object artefacts that immediately convey an unnatural impression to the viewer.  
Some works tackle the static grasp synthesis task using an explicit hand model \cite{krug2010efficient,grady2021contactopt, GRAB:2020} or an implicit representation \cite{graspingField2020}. 
However, considering the static frame alone is not sufficient to integrate the method into real-world applications such as AR/VR as it lacks information of the inherent scene dynamics. Recently, several works have been proposed to synthesize the hand and object interactions as a continuous sequence \cite{zhang2021manipnet,zhou2022toch,christen2022d}.
However, none of the state-of-the-art work explicitly considers an object's mass when generating hand-object interactions.
Real-life object manipulation, however, is substantially influenced by the mass of the objects we are interacting with. For example, we tend to grab light objects using our fingertips, whereas with heavy objects oftentimes the entire palm is in contact with the object. Manually creating such animations is tedious work requiring artistic skills. In this work, we propose  \method, \textit{i.e.,} the first learning-based \textit{mass conditioned} object manipulation synthesis method. The generated object manipulation naturally adopts its behavior depending on the object mass value. 
\method\ can synthesize such mass conditioned interactions given a trajectory plus action label (\textit{e.g.,} throw or move). 
The trajectory itself may also be generated conditioned on the action label and mass using the proposed cascaded diffusion model, or alternatively manually specified.
 
Specifically, given the action label and mass value as conditions, our cascaded diffusion model synthesizes the object trajectories as the first step. The synthesized object trajectory and mass value further condition a second diffusion model that synthesizes 3D hand motions and hand contact labels. After the final optimization step, \method\ returns diverse and physically plausible object manipulation animations. We also demonstrate a simple but effective data capture set-up to produce a 3D object manipulation dataset with corresponding mass values. The contributions of our work are as follows:
\begin{itemize} 
\itemsep0em 
\item The first approach to synthesize \textit{mass-conditioned} object manipulations in 3D. 
Our setting includes two hands and a single object of varying mass. 
\item A cascaded denoising diffusion model for generating trajectories of hands and objects allowing different types of conditioning inputs.
Our approach can both synthesize new object trajectories and operate on user-provided trajectories (in this case, the object trajectory synthesis part is skipped). 
\item A new component for introducing plausible dynamics into user-provided trajectories. 

\end{itemize} 

Our experiments confirm that \method\ synthesizes qualitatively and quantitatively more plausible 3D object manipulations compared with other baselines. 
\method\ shows plausible manipulative interactions even for mass values vastly different from those seen during the training. 

%% file: sec/2_related_works.tex
\section{Related Work}
There has been a significant amount of research in the field of 3D hand-object interaction motion synthesis. Here, we will review some of the most relevant works in this area.  Grasp synthesis works are discussed in Sec.~\ref{ssec:grasph_synthesis} and works that generate hand-object manipulation sequences in Sec.~\ref{ssec:obj_manip}. Lastly, closely related recent diffusion model based synthesis approaches are discussed in Sec.~\ref{ssec:diffusion_syn}.

\subsection{Grasp Synthesis}\label{ssec:grasph_synthesis} Synthesising physically plausible and natural grasps bears a lot of potential downstream applications. Thus, many works in this field have been proposed in computer graphics and vision \cite{pollard2005physically,li2007data,zhang2021manipnet,ghosh2022imos,ye2012synthesis}, and robotics community \cite{thobbi2010imitation,krug2010efficient}. ContactOpt \cite{grady2021contactopt} utilizes a differentiable contact model to obtain a plausible grasp from a hand and object mesh. Karunratanakul \etal \cite{graspingField2020} proposed a \textit{grasping field} for a grasp synthesis where hand and object surfaces are implicitly represented using a signed distance field.
Zhou \etal \cite{zhou2022toch} proposed a learning-based object grasp refinement method given noisy hand grasping poses. GOAL \cite{taheri2022goal} synthesizes a whole human body motion with grasps along with plausible head directions. These works synthesize natural hand grasp on a variety of objects. However, unlike the methods in this class, we synthesize a sequential object manipulation, changing not only the hand pose but also object positions bearing plausible hand-object interactions. %

\subsection{Object Manipulation}\label{ssec:obj_manip} Synthesising a sequence for object manipulation is challenging since the synthesized motions have to contain temporal consistency and plausible dynamics considering the continuous interactions. %
Ghosh \etal \cite{ghosh2022imos} proposed a human-object interaction synthesis algorithm associating the intentions and text inputs. ManipNet \cite{zhang2021manipnet} predicts dexterous object manipulations with one/two hands given $6$ DoF of hands and object trajectory from a motion tracker. CAMS \cite{Zheng_2023_CVPR} synthesizes hand articulations given a sequence of interacting object positions. Unlike these approaches, our algorithm \textbf{synthesizes} the $6$ DoF of the hands and objects as well as the finger articulations affected by the conditioned mass values. D-Grasp \cite{christen2022d} is a reinforcement learning-based method that leverages a physics simulation to synthesize a dynamic grasping motion that consists of approaching, grasping and moving a target object. In contrast to D-Grasp, our method consists of a cascaded diffusion model architecture and has explicit control over the object mass value that influences the synthesized interactions. Furthermore, D-Grasp uses a predetermined target grasp pose and therefore does not faithfully adjust its grasp based on the mass value in the simulator unlike ours.

\subsection{Diffusion Model based Synthesis}\label{ssec:diffusion_syn}
Recently, diffusion model \cite{sohl2015deep} based synthesis approaches have been receiving growing attention due to their promising results in a variety of research fields \eg image generation tasks \cite{saharia2022photorealistic,rombach2022high,ho2020denoising}, audio synthesis \cite{kong2020diffwave}, motion synthesis \cite{yuan2022physdiff,zhang2022motiondiffuse,tevet2022human,dabral2022mofusion} and 3D character generation from texts \cite{poole2022dreamfusion}. MDM \cite{tevet2022human} shows the 3D human motion synthesis and inpainting tasks from conditional action or text inputs utilizing a transformer-based architecture allowing the integration of the geometric loss terms during the training. Our method is the first diffusion model based approach that synthesizes hand-object interactions. Furthermore, unlike the existing works in the literature, we condition the synthesized motions on a physical property, \textit{i.e.,} object mass. 

%% file: sec/3_method.tex
\section{Method}
\label{sec:method}

\begin{figure*}[t!] 
\includegraphics[width=\linewidth]{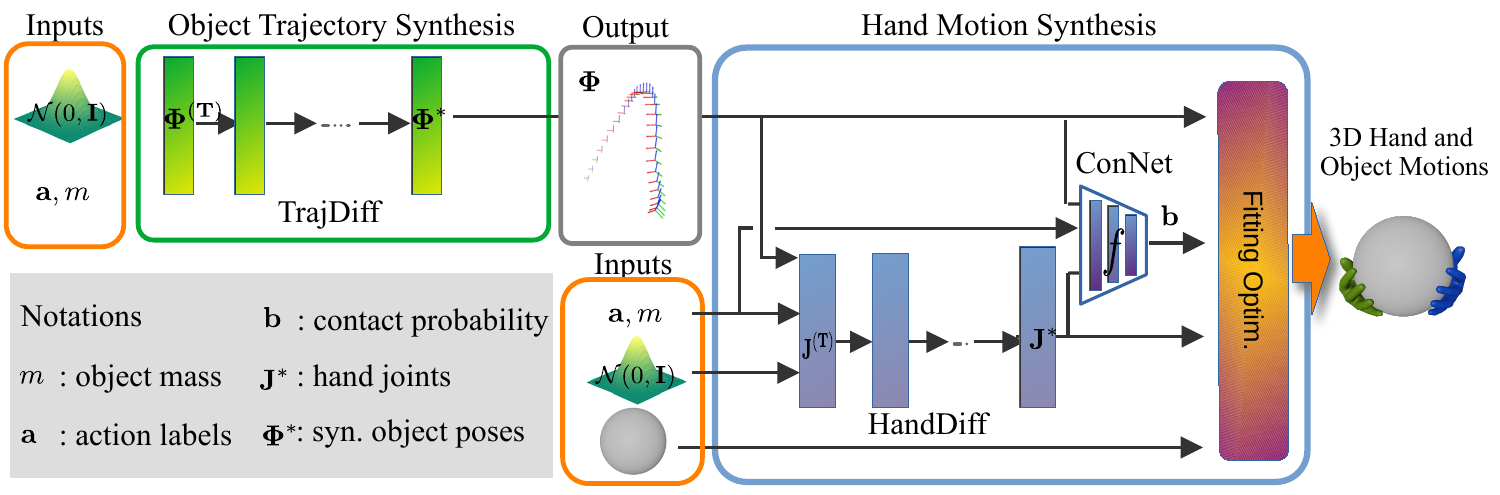} 
\caption{\textbf{The proposed framework.} 
The object trajectory synthesis stage accepts as input the conditional mass value $m$ and action label $\mathbf{a}$ along with a Gaussian noise sampled from $ \mathcal{N}(0,\mathbf{I})$, and outputs an object trajectory. The hand motion synthesis stage accepts $\mathbf{a}$, $m$ and the synthesized trajectory as conditions along with a gaussian noise sampled from  $ \mathcal{N}(0,\mathbf{I})$. ConNet in this stage estimates the per-vertex hand contacts from the synthesized hand joints, object trajectory and conditioning values $\mathbf{a}$, $m$. The final fitting optimization step returns a set of 3D hand meshes that plausibly interact with the target object. } \label{fig:overview}
\end{figure*}

Our goal is to synthesize 3D motion sequences of two hands interacting with an object whose mass affects both the trajectory of the object and the way the hands grasp it. %
The inputs of this method are a conditional scalar mass value and optionally a one-hot coded action label and/or a manually drawn object trajectory. Our method synthesizes a \textit{motion} represented as $N$ successive pairs of 3D hands and object poses.
To this end, we employ denoising diffusion models (DDM) \cite{sohl2015deep} 
for 3D hand motion and object trajectory synthesis; see Fig.~\ref{fig:overview} for the overview. 
We first describe our mathematical modeling and assumptions in Sec.~\ref{ssec:modelling}. In Secs.~\ref{ssec:hand_syn} and \ref{ssec:traj_syn}, we provide details of our hand motion synthesis network \textit{HandDiff} and trajectory synthesis algorithm \textit{TrajDiff}, respectively. We describe the method to synthesize the 3D motions given user input trajectory in Sec.~\ref{sssec:traj_draw}. The details of network architectures and training are elaborated in our supplementary material. 

\subsection{Assumptions, Modelling and  Preliminaries}\label{ssec:modelling} 

In this work, we assume that the target object is represented as a mesh. 3D hands are represented with a consistent topology, which is described in the following paragraph. 

\paragraph{Hand and Object Modelling} To represent 3D hands, we employ the hand model from GHUM \cite{xu2020ghum} which is a nonlinear parametric model learned from large-scale 3D human scans. %
The hand model from GHUM defines the 3D hand mesh as a differentiable function $\mathcal{M}(\btau, \bphi,\btheta,\bbeta)$ of global root translation $\btau\,{\in}\,\mathbb{R}^{3}$, global root orientation $\bphi\,{\in}\,\mathbb{R}^{6}$ represented in 6D rotation representation \cite{zhou2019continuity}, pose parameters $\btheta\,{\in}\,\mathbb{R}^{90}$   and shape parameters $\bbeta\,{\in}\,\mathbb{R}^{16}$. We employ two GHUM hand models to represent left and right hands, which return hand vertices $\bv\,{\in}\,\mathbb{R}^{3l}$ ($l=1882=941\cdot2$) and 3D hand joints $\bj\,{\in}\,\mathbb{R}^{3K}$ ($K=42=21\cdot2$). The object pose is represented by its 3D  translation $\btau_{\text{obj.}} \,{\in}\,\mathbb{R}^{3}$ and rotation $\bphi_{\text{obj.}}\,{\in}\,\mathbb{R}^{6}$. 
Our method \method\ synthesizes $N$ successive (i) 3D hand motions represented by the hand vertices $\bV \,{=}\,\{\bv_{1},...,\bv_{N}\}\,{\in}\,\mathbb{R}^{N\times3l}$ and hand joints $\bJ \,{=}\,\{\bj_{1},...,\bj_{N}\}\,{\in}\,\mathbb{R}^{N\times3K}$, and (ii) optionally object poses 
\begin{equation} 
\bPhi \,{=}\,\{\bPhi_{1},...,\bPhi_{N}\}\,{\in}\,\mathbb{R}^{N\times (3+6)}, 
\end{equation} 
where $\bPhi_{i}= [\btau_{\text{obj.},i},\bphi_{\text{obj.},i}]$. 
The object pose is defined in a fixed world frame $\mathcal{F}$, and the global hand translations are represented relative to the object center position. The global hand rotations are represented relative to $\mathcal{F}$.
\paragraph{Denoising Diffusion Model} The recently proposed Denoising Diffusion
 Probabilistic Model (DDPM) \cite{ho2020denoising} has shown compelling results both in image synthesis tasks and in motion generation tasks \cite{tevet2022human}. Compared to other existing generative models (\textit{e.g.,} VAE \cite{sohn2015learning} or GAN \cite{NIPS2014_5ca3e9b1}) that are often employed for motion synthesis tasks, the training of DDPM is simple, as it is not subject to the notorious mode collapse while generating motions of high quality and diversity.

Following the formulation by \citet{ho2020denoising}, the forward diffusion process is defined as a Markov process adding Gaussian noise in each step. 
The noise injection is repeated $T$ times. 
Next, let $\mathbf{X}^{(0)}$ be the original ground-truth (\GT{}) data (without noise). 
Then, the forward diffusion process is defined by a distribution $q(\cdot)$: 
\begin{equation}\label{eq:forward_diffusion}
q\left(\mathbf{X}^{(1:T)}\mid \mathbf{X}^{(0)}\right)=\prod_{t=1}^T q\left(\mathbf{X}^{(t)} \mid \mathbf{X}^{(t-1)}\right),
\end{equation}

{
\small
\begin{equation}\label{eq:forward_diffusion2}
q\left(\mathbf{X}^{(t)} \mid \mathbf{X}^{(t-1)} \right)=\mathcal{N}\left(\mathbf{X}^{(t)}\mid\sqrt{1-\beta_t} \mathbf{X}^{(t-1)}, \beta_{t} \mathbf{I}\right),
\end{equation} 
}
where $\beta_{t}$ are constant hyperparameters (scalars) that are fixed per each diffusion time step $t$. 
Using a reparametrization technique, we can sample $\mathbf{X}^{(t)}$ using the original data $\mathbf{X}^{(0)}$ and standard Gaussian noise $\epsilon{\sim}\mathcal{N}(0,I)$: 
\begin{equation}\label{eq:reparam}
\mathbf{X}^{(t)} = \sqrt{\alpha_{t}}\mathbf{X}^{(0)}+ \sqrt{1-\alpha_t}\epsilon,
\end{equation} where $\alpha_{t} = \prod_{i=1}^t (1-\beta_{i})$. The network is trained to reverse this process by denoising on each diffusion time step starting from a standard normal distribution $\mathbf{X}^{(T)}{\sim}\mathcal{N}(0,I)$:
\begin{equation}\label{eq:reverse_diffusion}
p\left(\mathbf{X}^{(0:T)}\right)=p\left(\mathbf{X}^{(T)}\right) \prod_{t=1}^T p\left(\mathbf{X}^{(t-1)} \mid \mathbf{X}^{(t)}\right),
\end{equation} 
where $p\left(\mathbf{X}^{(t-1)} \mid \mathbf{X}^{(t)}\right)$ denotes the conditional probability distribution estimated from the network output. From Eq.~\eqref{eq:reverse_diffusion}, we obtain the meaningful generated result $\mathbf{X}^*$ after $T$ times of denoising process. that follows the data distribution of the training dataset.
In the formulation of DDPM \cite{ho2020denoising}, the network is trained to predict the added noises on the data for the reverse diffusion process. The \textit{simple} loss term  is formulated as
\begin{equation}\label{eq:simple}
\mathcal{L}_{\text {simple}}=E_{\epsilon, t \sim[1, T]}\left[\left\|\epsilon-\epsilon_{\theta}\left(\mathbf{X}^{(t)}, t, c\right)\right\|_2^2\right], 
\end{equation}
where $c$ denotes an optional conditioning vector. 
 The loss term of Eq.~\eqref{eq:simple} drives the network $\epsilon_{\theta}$ towards predicting the added noise. Training the network with Eq.~\eqref{eq:simple} alone already generates highly diverse motions. 

In our case $\mathbf{X}^{*}$ represents sequences of 3D points corresponding to the synthesized motion trajectories (for hands and objects). 
Unfortunately, Eq.~\eqref{eq:simple} alone often leads to artifacts in the generated sequences such as joint jitters and varying bone length when applied to motion synthesis. 
To improve the plausibility of the generated results, \citet{dabral2022mofusion} proposed an algorithm to integrate the explicit geometric loss terms into the training of DDPM.
At an arbitrary diffusion time step $t$, we can obtain the approximated original data $\hat{\mathbf{X}}^{(0)}$ using the estimated noise from $\epsilon_{\theta}$  instead of $\epsilon$ in Eq.~\eqref{eq:reparam} and solving for $\hat{\mathbf{X}}^{(0)}$:
\begin{equation}\label{eq:sampling}
\hat{\mathbf{X}}^{(0)}=\frac{1}{\sqrt{\alpha}} \mathbf{X}^{(t)}-\left(\sqrt{\frac{1}{\alpha}-1}\right) \epsilon_\theta\left(\mathbf{X}^{(t)}, t, c\right).
\end{equation} 
During the training, geometric penalties can be applied on $\hat{\mathbf{X}}^{(0)}$ so as to prevent the aforementioned artifacts.
In the following sections, we follow the mathematical notations of DDPM literature \cite{ho2020denoising,dabral2022mofusion} as much as possible. The approximated set of hand joints and object poses obtained from Eq.~\eqref{eq:sampling} are denoted $\hat{\bJ}^{(0)}$ and $\hat{\bPhi}^{(0)}$, respectively. Similarly, the synthesized set of meaningful hand joints and object poses obtained from the reverse diffusion process Eq.~\eqref{eq:reverse_diffusion} are denoted  $\bJ^{*}$ and $\bPhi^{*}$, respectively.

\subsection{Hand 3D Motion Synthesis}\label{ssec:hand_syn}  
Our DDPM-based architectures \textit{HandDiff} $\mathcal{H}(\cdot)$ and \textit{TrajDiff} $\mathcal{T}(\cdot)$ are based on the stable diffusion architecture \cite{rombach2022high} with simple 1D and 2D convolution layers (see our supplementary for more details). 
During the training, we follow the formulation of \citet{dabral2022mofusion} described in Sec.~\ref{ssec:modelling} to introduce geometric penalties on $\hat{\bJ}^{(0)}\,{\in}\,\mathbb{R}^{N\times3K}$ and $\hat{\bPhi}^{(0)}\,{\in}\,\mathbb{R}^{N\times 9}$ combined with the simple loss described in Eq.~\eqref{eq:simple}.  
\paragraph{Hand Keypoints Synthesis} In this stage, we synthesize a set of 3D hand joints and per-vertex hand contact probabilities. Knowing the contact positions on hands substantially helps to reduce the implausible "floating object" artifacts of the object manipulation (see Sec.\ref{sec:experiments} for the ablations). The synthesized 3D hand joints and contact information are further sent to the final fitting optimization stage where we obtain the final hand meshes considering the plausible interactions between the hands and the object.  
 
Our diffusion model based \textit{HandDiff} $\mathcal{H}(\cdot)$  accepts as inputs a 3D trajectory $\bPhi\,{\in}\, \mathbb{R}^{N\times (3+6)}$ and mass scalar value $m$ where $N$ is the number of frames of the sequence. From the reverse diffusion process of $\mathcal{H}(\cdot)$, we obtain the synthesized set of 3D joints $\bJ^{*}\,{\in}\, \mathbb{R}^{N\times 3K}$. $\bPhi$ can be either  synthesized by \textit{TrajDiff} $\mathcal{T}(\cdot)$ (Sec.~\ref{ssec:traj_syn_by_Net}) or manually provided (Sec.~\ref{sssec:traj_draw}).  

Along with the set of 3D hand joint positions, our 1D convolution-based \textit{ConNet} $f(\cdot)$  also estimates the contact probabilities $\mathbf{b}\,{\in}\, \mathbb{R}^{N\times l}$ on the hand vertices from the hand joint and object pose sequence with a conditioning vector $\bc$ that consists of a mass value $m$ and an action label $\mathbf{a}$. %

\textit{ConNet} $f(\cdot)$ is trained using a binary cross entropy ($\operatorname{BCE}$) loss with the \GT{} hand contact labels $l_{\text{con.}}$:
\begin{equation} \label{eq:hand_con} 
\mathcal{L}_{\text{con.}}=   \operatorname{BCE}( f(\bJ^{(0)},\bPhi^{(0)},\bc), l_{\text{con.}}),
\end{equation} where $\bJ^{(0)}$ and $\bPhi^{(0)}$ denotes a set of \GT{} 3D hand joints and \GT{} object poses, respectively.
At test time, \textit{ConNet} estimates the contact probabilities from the synthesized 3D hand joints and object positions conditioned on $\bc$. The estimated contact probabilities $\mathbf{b}$ are used in the subsequent \textit{fitting optimization} step, to increase the plausibility of the hand and object interactions. 

The objective $\mathcal{L}_{\text{H}}$ for the training of \textit{HandDiff} reads: 
\begin{equation} \label{eq:handdif} 
\mathcal{L}_{\text{H}} =  \mathcal{L}_{\text{simple}}
+\lambda_{\text{geo}}\mathcal{L}_{\text{geo}},
\end{equation}
where $\mathcal{L}_{\text{simple}}$ is computed following Eq.~\eqref{eq:simple} and

\begin{equation} \label{eq:handdif_geo} 
\mathcal{L}_{\text{geo}} =   
\lambda_{\text{rec.}} \mathcal{L}_{\text{rec.}}
+\lambda_{\text{vel.}} \mathcal{L}_{\text{vel.}}
+\lambda_{\text{acc}} \mathcal{L}_{\text{acc.}}
+\lambda_{\text{blen}} \mathcal{L}_{\text{blen.}} .
\end{equation}
$\mathcal{L}_{\text{rec.}}$, $\mathcal{L}_{\text{vel.}}$ and $\mathcal{L}_{\text{acc.}}$ are loss terms to penalize the positions, velocities, and accelerations of the synthesized hand joints, respectively:
\begin{equation} \label{eq:hand_rec} 
\mathcal{L}_{\text{rec.}}= \|\hat{\bJ}^{(0)} - \bJ^{(0)}\|^2_2,
\end{equation}

\begin{equation} \label{eq:hand_vel} 
\mathcal{L}_{\text{vel.}}= \|\hat{\bJ}^{(0)}_{\text{vel.}} - \bJ^{(0)}_{\text{vel.}}\|^2_2,
\end{equation}

\begin{equation} \label{eq:hand_acc} 
\mathcal{L}_{\text{acc.}}= \|\hat{\bJ}^{(0)}_{\text{acc.}} - \bJ^{(0)}_{\text{acc.}}\|^2_2, 
\end{equation}
where $\hat{\bJ}^{(0)}$ is an approximated set of hand joints from Eq.~\eqref{eq:sampling} and $\bJ^{(0)}$ denotes a set of GT hand joints. $\hat{\bJ}^{(0)}$ and $\bJ^{0}$ with the subscripts ``vel.'' and ``acc.'' represent the velocities and accelerations computed from their positions, respectively. 

$\mathcal{L}_{\text{blen.}}$ penalizes incorrect bone lengths of the hand joints using the function $d_{\text{blen}}: \mathbb{R}^{N\times 3K} \rightarrow \mathbb{R}^{N\times K}$ that computes bone lengths of hands given a sequence 3D hand joints of $N$ frames:
\begin{equation} \label{eq:hand_blen} 
\mathcal{L}_{\text{blen.}}=\|d_{\text{blen}}(\hat{\bJ}^{(0)}) - d_{\text{blen}}(\bJ^{(0)})\|^2_2.
\end{equation}
At test time, we obtain a set of 3D hand joints $\bJ^{*}$ using the denoising process detailed in Eq.~\eqref{eq:reverse_diffusion} given a Gaussian noise $\sim N\left(0, \mathbf{I}\right)$.%

\paragraph{Fitting Optimization} Once the 3D hand joint sequence $\bJ^{*}$ is synthesized from the trained $\mathcal{H}$, we solve an optimization problem to fit GHUM hand models to  $\bJ^{*}$. We use a threshold of $\mathbf{b}>0.5$ to select the effective contacts from the per-vertex contact probability obtained in the previous step.
Let $\mathbf{b}^{n}_{\text{idx}}\,{\subset}\,\llbracket 1,L \rrbracket$ be the subset of hand vertex indices with effective contacts on the $n$-th frame. The objectives are written as follows:
\begin{align} \label{eq:fittingOptim}
\small
\underset{\btau,\bphi,\btheta }{\operatorname{argmin} }(\lambda_{\text{data}} \mathcal{L}_{\text{data}}\!+\!\lambda_{\text{touch}} \mathcal{L}_{\text{touch}}\!+\!\lambda_{\text{col.}} \mathcal{L}_{\text{col.}} 
\!+\!\lambda_{\text{prior}} \mathcal{L}_{\text{prior}} ). %
\end{align}  
$\mathcal{L}_{\text{data}}$ is a data term to minimize the Euclidean distances between the GHUM ($\bJ$) and the synthesized hand joint key points ($\bJ^{*}$): 
\begin{equation} \label{eq:data_term} 
\mathcal{L}_{\text{data}} = \|\bJ -  \bJ^{*}\|^2_2.
\end{equation}
$\mathcal{L}_{\text{touch}}$ is composed of two terms. 
The first term reduces the distances between the contact hand vertices and their nearest vertices $\bP$ on the object to improve the plausibility of the interactions. 
The second term takes into account the normals of the object and hands which also enhances the naturalness of the grasp by minimizing the cosine similarity $s(\cdot)$ between the normals of the contact hand vertices $\mathbf{n}$ and the normals of their nearest vertices of the object  $\hat{\mathbf{n}}$.
{\small
\begin{align} \label{eq:touch_term} 
\mathcal{L}_{\text{touch}}\!=\!\sum_{i=1}^{N}\!\sum_{j\in\mathbf{b}^{i}_{\text{idx}}}\!\left\|\bV^{j}_i-\bP^{j}_{i}\right\|^{2}_{2}\!\!+\!\!\sum_{i=1}^{N}\!\sum_{i\in\mathbf{b}_{\text{idx}}}(1-s(\mathbf{n}^{j}_{i},\hat{\mathbf{n}}^{j}_{i})),
\end{align}}where the subscript $i$ denotes $i$-th sequence frame and the superscript $j$ denotes the index of the vertex with the effective contact.  
$\mathcal{L}_{\text{col.}}$ reduces the collisions between the hand and object by minimizing the penetration distances. Let $\mathcal{P}^{n}\,{\subset}\,\llbracket 1,U \rrbracket$ be the subset of hand vertex indices with collisions on $n$-th frame. Then we define
\begin{equation} \label{eq:col_term} 
\mathcal{L}_{\text{col.}} =\sum_{i=1}^{N} \sum_{j\in\mathcal{P}^{n}} \left\|\bV^{j}_{i}-\bP^{j}_{i}\right\|^{2}_{2}.
\end{equation}

$\mathcal{L}_{\text{prior}}$ is a hand pose prior term that encourages the plausibility of the GHUM hand pose by minimising the pose vector $\btheta$ of the GHUM parametric model 
\begin{equation} \label{eq:prior_term} 
\mathcal{L}_{\text{prior}} = \|\btheta\|^2_2.
\end{equation}
With all these loss terms combined, our final output shows a highly plausible hand and object interaction sequence. The effectiveness of the loss terms is shown in our ablative study (Sec.~\ref{ssec:quantitative}). Note that only for the non-spherical objects, which were not present in the training dataset, we apply a Gaussian smoothing on the hand and object vertices along the temporal direction with a sigma value~of~$3$ after the fitting optimization to obtain a smoother motion. %

\subsection{Object Trajectory Generation}\label{ssec:traj_syn} The input object trajectory for \textit{HandDiff} can be provided in two ways, (1) synthesizing 3D trajectory by \textit{TrajDiff} (Sec.\ref{ssec:traj_syn_by_Net}) or (2) providing a manual trajectory (Sec.~\ref{sssec:traj_draw}). The former allows generating an arbitrary number of hands-object interaction motions conditioned on mass values and action labels, which can contribute to a large-scale dataset generation for machine learning applications. The latter allows for tighter control of the synthesized motions which are still conditioned on an object's mass value but restricted to the provided trajectory.

\subsubsection{Object Trajectory Synthesis}\label{ssec:traj_syn_by_Net}    
To provide a 3D object trajectory to \textit{HandDiff}, we introduce a diffusion model-based architecture \textit{TrajDiff} that synthesizes an object trajectory given a mass value $m$ and an action label $\ba\,{\in}\,\mathbb{R}^{6}$ encoded as a one-hot vector. We observed that directly synthesizing a set of object rotation values causes jitter artifacts. 
We hypothesize that this issue comes 
\setlength{\columnsep}{8pt}
\begin{wrapfigure}{r} {0.2\textwidth}
  \centering 
  \includegraphics[width=0.2\textwidth]{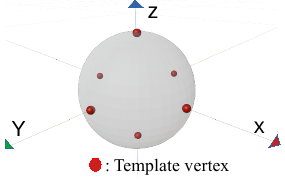}
  \vspace{-20pt}
  \caption{Definition of the template vertices.}
  \label{fig:template_vertices} 
  \vspace{-10pt}
\end{wrapfigure}
from simultaneously synthesizing two aspects of a pose, translation and rotation, each having a different representation. 
As a remedy, we propose to represent both the translation and rotation as 3D 
coordinates in a Cartesian coordinate system. 
Specifically, we first synthesize the \textit{reference vertex positions} $\bP_{\text{ref}}$ 
on the object surface defined in the object reference frame, and register them to the predefined \textit{template vertex positions} $\bP_{\text{temp}}$ to obtain the rotation of the object. 
We define $6$ template vertices as shown in Fig.~\ref{fig:template_vertices}. \textit{TrajDiff} thus synthesizes a set of reference vertex positions $\bP_{\text{ref}}\,{\in}\,\mathbb{R}^{N \times q}$ where $q = 18 (= 6 \times 3)$ that are defined in the object center frame along with a set of global translations. We then apply Procrustes alignment between $\bP_{\text{ref.}}$ and $\bP_{\text{temp.}}$ to obtain the object rotations. The objective of \textit{TrajDiff} is defined as follows: 
\begin{align} 
\label{eq:traj_diff} 
\mathcal{L}_{\mathcal{T}} =  \mathcal{L}_{\text{simple}}+ 
\lambda_{\text{geo.}}(&\lambda_{\text{rec.}} \mathcal{L}_{\text{rec.}}+\lambda_{\text{vel.}}\mathcal{L}_{\text{vel.}}\nonumber\\
+&\lambda_{\text{acc.}} \mathcal{L}_{\text{acc.}} +  \lambda_{\text{ref.}} \mathcal{L}_{\text{ref.}}). %
\end{align} 
 
$\mathcal{L}_{\text{rec.}}$, $\mathcal{L}_{\text{vel.}}$ and $\mathcal{L}_{\text{acc.}}$ follow the definitions given in Eqs.~\eqref{eq:hand_rec}, \eqref{eq:hand_vel} and \eqref{eq:hand_acc}, where $\bJ^{(0)}$ is replaced with \GT{} 3D object poses whose rotation is represented by the reference vertex positions instead of 6D rotation. 
$\mathcal{L}_{\text{ref}}$ is defined as: 
\begin{equation}
    \mathcal{L}_{\text{ref}} =  \|\hat{\bP}^{(0)}_{\text{ref}}  - \bP^{(0)}_{\text{ref}} \|^2_2 +  \|d_{\text{rel}}(\hat{\bP}^{(0)}_{\text{ref}} ) - d_{\text{rel}}(\bP^{(0)}_{\text{ref}})\|^2_2.
\end{equation} 
The first term of $\mathcal{L}_{\text{ref}}$ penalizes the Euclidean distances between the approximated reference vertex positions $\hat{\bP}^{(0)}_{\text{ref}}$ of Eq.~\eqref{eq:sampling} and the \GT{} reference vertex positions $\bP^{(0)}_{\text{ref}}$. The second term of $\mathcal{L}_{\text{ref}}$ penalizes the incorrect Euclidean distances of the approximated reference vertex positions relative to each other. To this end, we use a function $d_{\text{rel}}: \mathbb{R}^{N\times 3q} \rightarrow \mathbb{R}^{N\times q'}$, where $q'=\binom{q}{2}$, which computes the distances between all the input vertices pairs on each frame.  

The generated object trajectory responds to the specified masses. Thus, the motion range and the velocity of the object tend to be larger for smaller masses. In contrast, with a heavier object the trajectory shows slower motion and a more regulated motion range. 

\subsubsection{User-Provided Object Trajectory} 
\label{sssec:traj_draw} 
Giving the user control over the output in synthesis tasks is crucial for downstream applications such as character animations or avatar generation. Thanks to the design of our architecture that synthesizes 3D hand motions and hand contacts from a mass value and an object trajectory, a manually drawn object trajectory can also be provided to our framework as an input. 
However, manually drawing an input 3D trajectory is not straightforward, as it must consider the object dynamics influenced by the mass. For instance, heavy objects will accelerate and/or decelerate much slower than lighter ones.
Drawing such trajectories is tedious and often requires professional manual labour. 
To tackle this issue, we 
introduce a module that accepts 
a (user-specified) trajectory with an arbitrary number of points along with the object's mass, and outputs a \textit{normalized target trajectory (NTT)}. 

NTT is calculated from the input trajectory based on the intermediate representation that we call \textit{vector of ratios}, see our supplementary for its overview. First, the input (user-specified) trajectory is re-sampled uniformly to $N_{fix} = 720$ points and passed to \textit{RatioNet}, which for each time step estimates the distance traveled along the trajectory normalized to the range $[0, 1]$ (\eg the value of $0.3$ means that the object traveled $30\%$ of the full trajectory within the given time step). The NTT from this stage is further sent to the \textit{Hand Motion Synthesis} stage to obtain the final hand and object interaction motions. We next explain 1) the initial uniform trajectory re-sampling and 2) the intermediate ratio updates.

\noindent \textit{Uniform Input Trajectory Re-sampling.} 
To abstract away the variability of the number of points in the user-provided trajectory of $N_{\text{user}}$ points, we first interpolate it into a path $\Phi_{\text{fix}}$ of length $N_{\text{fix}}$ points. 
Note that $N_{\text{user}}$ is not fixed and can vary. We also compute the total path length  $d_{\text{user}}$ that is used as one of the inputs to the RatioNet network (elaborated in the next paragraph): 
\begin{equation}
    d_{\text{user}} = \sum_{i=1}^{N_{\text{fix}}-1} \|\bPhi^{i}_{\text{fix}}-\bPhi_{\text{fix}}^{i+1}\|^2,
\end{equation}
where $\bPhi_{\text{fix}}^{i}$ denotes the $i$-th object position in $\bPhi_{\text{fix}}$. 

\noindent\textit{Intermediate Ratio Updates Estimation.} 
From the normalized object path $\Phi_{\text{fix}}$, a total distance of the path $d_{\text{user}}$, and mass $m$, we obtain the information of the object location in each time step using a learning-based approach. To this end, we introduce a MLP-based network \textit{RatioNet} $R(\cdot)$ that estimates the location of the object along the path $\Phi_{\text{fix}}$
encoded as a ratio starting from the beginning, 
see our supplementary for the schematic visualization. Specifically, \textit{RatioNet} accepts the residual of $\Phi_{\text{fix}}$ denoted as $\bar{\Phi}_{\text{fix}}$, a mass scalar value and $d_{\text{user}}$ and outputs a vector $\mathbf{r} \,{\in}\,\mathbb{R}^{N}$ that contains the update of the ratios on the path for each time step: 
\begin{equation}
    \mathbf{r}= R(\bar{\Phi}_{\text{fix}}, m, d_{\text{user}}).
\end{equation}
Next, we obtain the cumulative ratios $\mathbf{r}_{cuml}$ from $\mathbf{r}$ starting from the time step $0$ to the end of the frame sequence. Finally, the NTT $\mathbf{\Phi}_{\text{NTT}} = [\mathbf{\Phi}_{\text{NTT}}^{0},...,\mathbf{\Phi}_{\text{NTT}}^{N}] $ at time step $t$ is obtained as: 
\begin{equation}
    \mathbf{\Phi}_{\text{NTT}}^{t} = \Phi_{\text{fix}}^{id},\;\,\text{with}\;\,
   id = \operatorname{round}(r_{cum}^{t} \cdot N_{\text{fix}}), 
\end{equation} 
where $id$ and ``$\cdot$'' denote the index of $\Phi_{\text{fix}}$, and multiplication, respectively. \textit{RatioNet} is trained with the following loss function $\mathcal{L_{\text{ratio}}}$:
\begin{equation} \label{eq:ratio_loss} 
\mathcal{L_{\text{ratio}}}\!=\!\|\mathbf{r} -  \hat{\mathbf{r}}\|^2_2+\|\mathbf{r}_{\text{vel}} -   \hat{\mathbf{r}}_{\text{vel}}\|^2_2+\|\mathbf{r}_{\text{ac.}} -  \hat{\mathbf{r}}_{\text{acc}}\|^2_2 + \mathcal{L}_{one},
\end{equation}
\begin{equation} \label{eq:one_loss} 
\mathcal{L}_{one} = \|(\sum_{i=1}^{N} \mathbf{r}^{i})-1 \|^2_2,
\end{equation} where $\hat{\mathbf{r}}$ denotes the \GT{} ratio updates. 
Note that all terms in Eq.~\eqref{eq:ratio_loss} have the same weights. 
The subscripts ``vel.'' and ``acc.'' represent the velocity and accelerations of $\mathbf{r}$ and $\hat{\mathbf{r}}$, respectively. 
$\mathcal{L}_{one}$ encourages \textit{RatioNet} to estimate the sum of the ratio updates to be $1.0$.

%% file: sec/4_experiments.tex
\section{Experiments}\label{sec:experiments}
To the best of our knowledge, there exists no other work that addresses the hand object manipulation synthesis conditioned on mass. Therefore, we compare our method mainly with two baseline methods which, similarly to our method, employ an encoder-decoder architecture, but which are based on the popular methods VAE \cite{kingma2013auto} and VAEGAN \cite{yu2019vaegan}. Specifically, the VAE baseline uses the same diffusion model architecture as our method, but we add a reparameterization layer \cite{kingma2013auto} and remove the skip connections between the encoder and the decoder. The VAEGAN baseline shares the same architecture of the generator, while the discriminator network consists of three 1D convolution layers and two fully connected layers at the output of the network, and we use ELU activation in the discriminator \cite{clevert2015fast}. The generator and discriminator networks are conditioned by the same conditioning vector. In all the following experiments we will refer to our proposed method as \ours{} and to the baselines as \vae{} and \vaegan{}. We also compare with ManipNet \cite{zhang2021manipnet} qualitatively, while the quantitative comparison is omitted due to the following limitations of ManipNet. (1) It requires a sequence of 6D hand and object poses as inputs, whereas our approach only needs conditioning of mass value and an optional action label, (2) certain evaluation metrics (e.g., diversity, multimodality)  cannot be fairly computed on ManipNet due to its deterministic nature, and (3) ManipNet lacks control over the object weight as it does not support mass conditioning. Therefore, we compare qualitatively with ManipNet by inputting the ground truth 6D object and hand poses to the method. Please refer to our supplementary material for additional quantitative experiments (additional ablations, qualitative results, and a user study).

\subsection{Quantitative Results}\label{ssec:quantitative}
In this section, we evaluate the motion quality of \method\ from various perspectives. We report a diversity and multi-modality measurement as suggested by Guo \etal \cite{guo2020action2motion} in Table \ref{tab:diversity}. We also evaluate the physical plausibility by measuring the following metrics:\\
\textbf{Non-collision ratio ($\mcol$)} measures the ratio of frames with no hand-object collisions. A higher value indicates fewer collisions between the hand and the object.\\ 
\textbf{Collision distance ($\mdist$)} measures the distance of hand object penetration averaged over all the samples. A lower value indicates low magnitude of the collisions.\\
\textbf{Non-touching ratio ($\mtouch$)} measures the ratio of samples over all the samples where there is no contact between the hand and object. A lower value indicates fewer \textit{floating object} artifacts (\textit{i.e.,} spurious absence of contacts). 

Note that to report $\mtouch$, we discard throwing motion action labels, as the assumption is that there should be constant contact between the hands and the object. The hand vertices whose nearest distances to the object are lower than a threshold value of $5mm$ are considered contact vertices. Similarly, to compute $\mcol$ and $\mdist$, the interpenetrations over $5mm$ are considered collisions. To compute the metrics, we generate $500$ samples across $6$ different action labels.

\vspace{-12pt} 
\paragraph{Diversity and Multimodality} 
Diversity measures the motion variance over all the frames within each action class, whereas multimodality measures the motion variance across the action classes. High diversity and multimodality indicate that the generated samples contain diversified motions. Please refer to \citet{guo2020action2motion} for more details. We report the diversity and multimodality metrics for the generated hand motions and the object trajectories in Table \ref{tab:diversity}. It is clear that in both cases \ours{} generates much more diversified motions when compared to the baselines, which we attribute to our diffusion model-based architecture. Notably, the generated trajectory samples contain more diversified motions compared with the metrics computed on the \GT{} data.
\vspace{-12pt} 

\paragraph{Physical plausibility}
We report the physical plausibility measurements in Table \ref{tab:plausibility}. \ours{} shows the highest performance across all three metrics $\mcol$, $\mdist$ and $\mtouch$. \vae{} yields $\mcol$ and $\mdist$ comparable to \ours{}, however, its $\mtouch$ is substantially higher with $42\%$ error increase compared to \ours{}. \vaegan{} shows $\mtouch$ similar to \ours{} but it underperforms in terms of the collision-related metrics.

\begin{figure*}[t!] 
\includegraphics[ width=1\linewidth ]{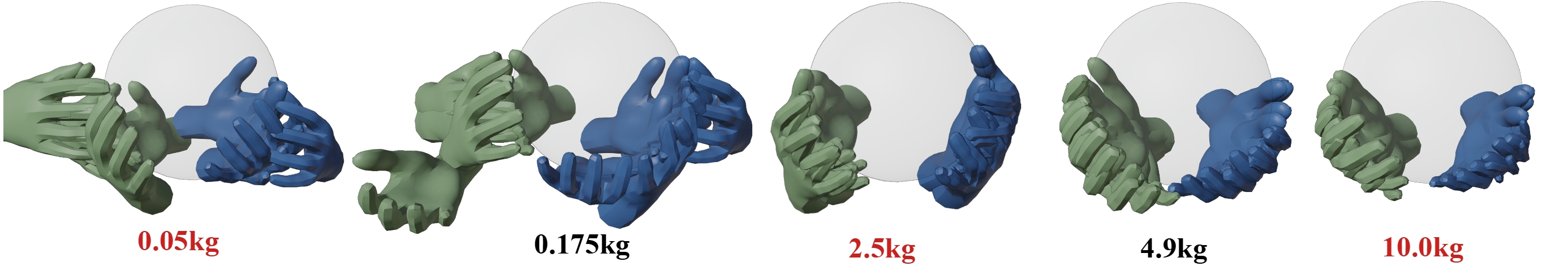} 
\caption{ 
Grasp synthesis with different object masses. 
Our method can generate sequences influenced by masses close (in black) and far (in red) from the training dataset. 
Note that in the case of small masses, hands can support the object with fingertips and release the object for some time; the hands are generally more mobile. 
The situation is different for moderate and large masses: A larger area supporting the object is necessary, and the hands are less mobile. 
} 
\vspace{-12pt} 
\label{fig:diff_mass_grasphs}
\end{figure*}   
\vspace{-12pt} 
\paragraph{Ablation study}
Here, we motivate the use of the important loss terms of our fitting optimization and training loss functions. In Table~\ref{tab:plausibility}, we show the results of the fitting optimization without $\mathcal{L}_{touch}$ and without $\mathcal{L}_{col.}$. When omitting the contact term $\mathcal{L}_{touch}$, the generated hands are not in contact with the object in most of the frames. This results in substantially higher metric $\mtouch$ and manifests through undesirable \textit{floating object} artifacts. 
Omitting the collision term $\mathcal{L}_{col.}$ leads to frequent interpenetrations, lower $\mcol$ and higher $\mdist$. Therefore, it is essential to employ both the loss terms to generate sequences with higher physical plausibility. 
For more ablations for the loss terms $\mathcal{L}_{vel.}$ and $\mathcal{L}_{acc.}$ for the network training, also for ablation on RatioNet, please refer to our supplementary material.

\subsection{Qualitative Results}\label{ssec:qualitative}

\paragraph{Hand-Object Interaction Synthesis}
In our supplementary video, we show the synthesized hand and object interaction sequence conditioned by the action labels and mass of the object. The synthesized motions show realistic and dynamic interactions between the hands and the object. Furthermore, thanks to our cascaded diffusion models, the generated motions show high diversity. The results thus visually clearly complement the quantitative findings listed in Table \ref{tab:diversity}. Furthermore, our method shows a more robust and plausible synthesis that faithfully responds to the conditioning mass value compared to ManipNet \cite{zhang2021manipnet}. 

 {
\renewcommand{\arraystretch}{1.2}
\begin{table}[t]  
\centering 
\scalebox{0.70}{ 
\begin{tabular} { ccccc  }\toprule
        & \multicolumn{2}{c}{Hand synthesis}& \multicolumn{2}{c}{Trajectory synthesis}\\
        \cmidrule(lr){2-3}\cmidrule(lr){4-5}
       &  Diversity $\uparrow$ & Multimodality $\uparrow$ &  Diversity $\uparrow$ & Multimodality $\uparrow$ \\   \midrule
     \GT{} & $9.984^{\pm 0.36}$ & $7.255^{\pm 0.32}$  & $10.041^{\pm 0.28}$ & $7.374^{\pm  0.29}$  \\ \midrule
     Ours & $\bf{9.606^{\pm 0.33}}$ & $\bf{7.07^{\pm 0.30}}$& $\bf{11.01^{\pm 0.37}}$ & $\bf{8.05^{\pm 0.33}}$ \\
     VAE   & $8.350^{\pm 0.42}$   & $6.0465^{\pm 0.34}$ & $9.584^{\pm 0.47}$  & $7.696^{\pm 0.43}$  \\
     VAEGAN &  $7.821^{\pm 0.27}$  &  $5.887^{\pm 0.26}$ & $8.428^{\pm 0.29}$  & $6.285^{\pm 0.30}$  \\\bottomrule
\end{tabular} }\caption{\label{tab:diversity} Diversity and multimodality 
for the hand and trajectory synthesis compared to the ground truth. } 
\end{table}
}

\paragraph{Grasp Synthesis}
We show 5 samples of grasps for different conditioning mass values in Fig. \ref{fig:diff_mass_grasphs}. To generate this visualization, we trained \textit{HandDiff} without providing the action labels. In order to synthesize the graphs, we provide an object trajectory with position and rotations set to $0$. Our method shows diverse grasps faithfully reflecting the conditional mass values. Most notably, the synthesized hands tend to support the heavy object at its bottom using the whole palm, whereas the light object tends to be supported using the fingertips only. Furthermore, the synthesized grasps show reasonable results even with unseen interpolated ($2.5$kg) and extrapolated ($0.05$kg and $10.0$kg) mass values (highlighted in red).%

\begin{table}[t]  
\centering 
\scalebox{0.92}{ 
\begin{tabular} { cccc  }\toprule
       &  $\mcol$ [$\%$] $\uparrow$ & $\mdist$ [mm] $\downarrow$& $\mtouch$ [$\%$] $\downarrow$ \\   \midrule
     Ours & \bf{97.84} & \bf{0.041} & \bf{1.97}  \\
     Ours w/o $\mathcal{L}_{\text{touch}}$&100.0  & 0.0  & 63.3 \\
     Ours w/o $\mathcal{L}_{\text{col.}}$& 38.41 & 0.296  & 1.88  \\
     VAE   & 97.2  & 0.055  & 2.80  \\
     VAE-GAN & 96.03  & 0.058 & 2.03 \\\bottomrule
\end{tabular}}
\caption{\label{tab:plausibility} Physical plausibility measurement of our full model and its trimmed versions \textit{vs} VAE and VAE-GAN.}
\end{table}
 

%% file: sec/5_conclusions.tex
\section{Conclusion} 
This paper introduces the first approach to synthesize realistic 3D object manipulations with two hands faithfully responding to conditional mass. Our diffusion-model-based \method\ approach produces plausible and diverse object manipulations, as verified quantitatively and qualitatively.%

Since this topic has so far been completely neglected in the literature, %
the focus of this paper is to demonstrate the impact of mass onto manipulation and hence we opted to use a single shape with uniform static mass distribution.
As such there are several limitations that open up to exciting future work; for example the effect of shape diversity, non-uniform mass distribution (i.e. one side of the object is heavier than the other), or dynamic mass distribution (\eg, a bottle of water). 
Furthermore, we would like to highlight that other physical factors, such as friction or individual muscle strength, also impact object manipulation and could be addressed in future works. Lastly, while this work focused on synthesis with applications for ML data generation, entertainment and mixed reality experiences, we believe that weight analysis is another interesting avenue to explore, i.e. predicting the weight based on observed manipulation. This could be valuable in supervision scenarios to identify if an object changed its weight over time. 

%% file: sec/6_appendix.tex
\appendix 
\clearpage
\noindent{\fontsize{12}{12}\selectfont \textbf{Appendices}}
\vspace{0.1cm}

This supplementary document provides the details of our dataset acquisition (Sec. ~\ref{sec:dataset}), network architectures (Sec. ~\ref{sec:netarch}), and implementations (Sec. ~\ref{sec:implementations}). We also provide further ablations (1) for the loss terms $\mathcal{L}_{vel.}$ and $\mathcal{L}_{acc.}$ for the network training, (2) for the mass conditioning and (3) for ablation on RatioNet and (4) user-study on the synthesized motions. We also show additional qualitative results for (5) the objects unseen during the training, (6) visualizations of the synthesized contacts and (7) the synthesized motions given a user-provided trajectory  (Sec.~\ref{sec:further_evaluations}).

\section{Dataset} 
\label{sec:dataset} 
\setlength{\columnsep}{10pt} 
\begin{wrapfigure}{r} {0.2\textwidth}
  \centering 
  \vspace{-13pt}
  \includegraphics[width=0.2\textwidth]{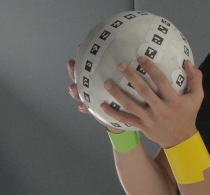}
  \vspace{-20pt}
  \caption{Image of our markered sphere and recording example.}
  \label{fig:capture_setup} 
  \vspace{-5pt}
\end{wrapfigure}
Since there exists no 3D hand and object interaction motion dataset with corresponding object mass values of the objects, we reconstruct such motions using $8$ synchronized Z-CAM E2 cameras of 4K resolution and $50$ fps. As target objects, we use five plastic spheres of the same radius $0.1$[m]. We fill them with different materials of different densities to prepare the objects of the same volume and different weights \ie $ 0.175, 2.0, 3.6, 3.9, 4.9$ kg. Each sphere is filled entirely so that its center of mass does not shift as the object is moved around. Five different subjects are asked to perform five different actions manipulating the object: (1) vertical throw and catch, (2) passing from one hand to another, (3) lifting up and down, (4) moving the object horizontally, and (5) drawing a circle. The subjects perform each action using both their hands while standing in front of the cameras and wearing colored wristbands (green for the right wrist and yellow for the left wrist), which are later used to classify handedness. The recordings from the multi-view setup were further used to reconstruct the 3D hand and object motions, totaling $110$k frames. The details of the capture and reconstruction processes are described in the following text. 
\begin{figure}[t!]  
\center
\includegraphics[width=0.775\linewidth]{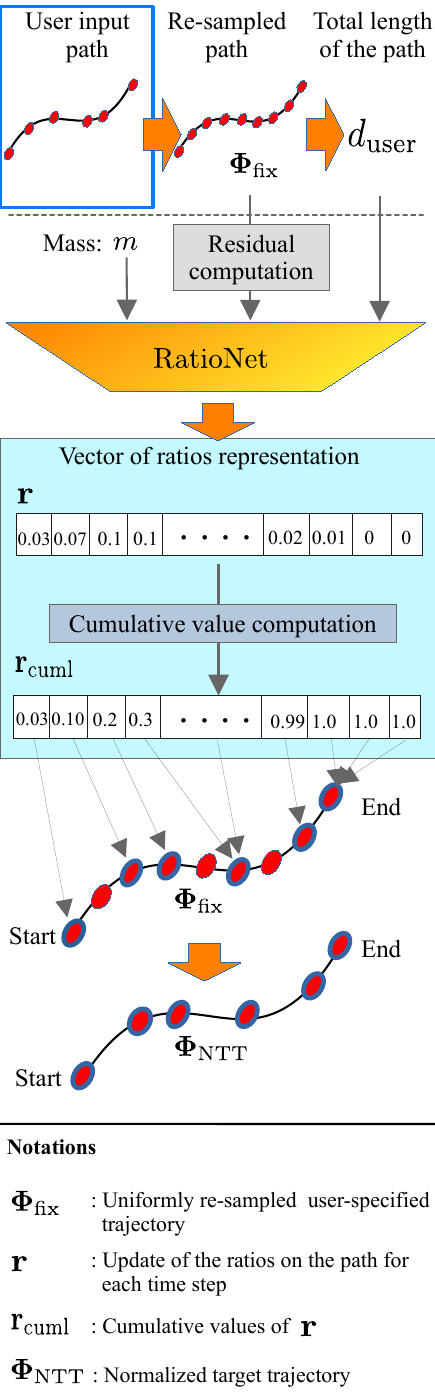} 
\caption{Schematic visualization of the user input trajectory processing stage.  
} \label{fig:schematic_draw}
\end{figure} 

\paragraph{Hand Motion Reconstruction} 
To reconstruct 3D hand motions, we first obtain 2D hand key points from all the camera views using MediaPipe \cite{lugaresi2019mediapipe}. We then fit GHUM hand models \cite{xu2020ghum} for both hands on each frame by solving 2D keypoint reprojection-based optimization with the known camera intrinsics and extrinsic combining with a collision loss term (Eq.(18)), a pose prior loss (Eq.(19)) in our main paper and a shape regularizer term that minimizes the norm of the shape parameter $\bbeta$ of the GHUM hand parametric model.

\paragraph{Object Trajectory Reconstruction}
We place around $50$ ArUco markers of the size $1.67\times1.67$ cm on each sphere for the tracking optimization (see Fig.~\ref{fig:capture_setup} for the example of our tracking object). The marker positions in the image space are tracked using the OpenCV \cite{opencv_library} library. The 3D object positions on each frame are obtained by solving the multi-view 2D reprojection-based optimization. %

\section{Network Architecture}\label{sec:netarch} 
We employ the Unet-based diffusion model networks from \citet{ho2020denoising} for our \textit{TrajDiff} and \textit{HandDiff}. \textit{HandDiff} uses four sets of 2D convolutional residual blocks for the encoder and decoder architecture. \textit{TrajDiff} is composed of two sets of residual blocks of 1D convolution layers instead of 2D convolutions. The number of kernels at its output 1D convolutional layer is set to $21$ which corresponds to the dimensionality of the object pose. \textit{ConNet} consists of three-1D convolutional layers with ELU and a sigmoid activation for its hidden layers and output layer, respectively. Similarly, \textit{RatioNet} is composed of three-layer MLP with ELU and a sigmoid activation functions in the hidden and output layers, respectively. 
Starting from the input layer, the output layer dimensions are $1024$, $512$ and $180$. See Fig.~\ref{fig:schematic_draw} for the overview of the user input trajectory processing stage (Sec. 3.3.2 in the main paper) that utilizes \textit{RatioNet}.

\begin{table}[t] 
 
\centering 
 \scalebox{1.05}{
\begin{tabular} { cccc  }\toprule
       &  Ours & Ours w/o $\mathcal{L}_{\text{vel}}$ & Ours w/o $\mathcal{L}_{\text{acc}}$\\   \midrule
     acc. dist. $\downarrow$& \bf{7.35}  & 26.4 &  11.2  \\ \bottomrule
\end{tabular}}\caption{\label{tab:accdist} Wasserstein distances between the acceleration distributions (``acc.~dist'') of the generated motions and ground-truth motions. Combining both $\mathcal{L}_{\text{vel}}$ and $\mathcal{L}_{\text{acc}}$ shows the highest plausibility in terms of the accelerations. } 
\end{table} 

 {
\renewcommand{\arraystretch}{1.0}
\begin{table}[t] 
 
\centering 
 \scalebox{0.78}{
\begin{tabular} { cccccc  }\toprule
    & $0.175$ [kg]&  $2.0$ [kg]& $3.6$ [kg] & $3.9$ [kg] & $4.9$ [kg] \\   \midrule
      ours &  \bf{0.006}   & \bf{0.010} & \bf{0.012} & \bf{0.011} & \bf{0.011}  \\ 
     ours w/o cond. & 0.089 &  0.070 &  0.081 & 0.061 & 0.074  \\\bottomrule 
\end{tabular}}\caption{\label{tab:accdist_conditioning} Wasserstein distances between the acceleration distributions (``acc.~dist'') of the generated and ground-truth motions.   } 
\end{table}
}

\section{Training and Implementation Details}\label{sec:implementations}
 
All the networks are implemented in TensorFlow \cite{tensorflow2015-whitepaper} and trained with $1$ GPU Nvidia Tesla V100 until convergence. The training of \textit{HandDiff}, \textit{TrajDiff}, \textit{ConNet} and \textit{RatioNet} takes $5$ hours, $3$ hours, $2$ hours and $2$ hours, respectively. We set the loss term weights of Eq.~(10) and (20) to $\lambda_{\text{rec.}}=1.0$, $\lambda_{\text{vel.}}=5.0$ and $\lambda_{\text{acc.}}=5.0$. $\lambda_{\text{blen.}}$ of Eq.~(10) and $\lambda_{\text{ref}}$ of Eq.~(20) are set to $10.0$ and $1.0$, respectively. For the fitting optimization defined in Eq.~(15), we set $\lambda_{\text{data}}=1.0$, $\lambda_{touch}=0.7$, $\lambda_{col.}=0.8$ and $\lambda_{\text{prior}}=0.001$. As in \citet{dabral2022mofusion}, $\lambda_{\text{geo.}}$ of Eq.~(10) and (20) are set such that larger penalties are applied with smaller diffusion steps $t$:
\begin{equation}
     \lambda_{\text{geo.}}  = \frac{10}{\exp{\frac{10t}{T}}},
\end{equation} where $T$ is the maximum diffusion step. We empirically set the maximum diffusion step $T$ for \textit{HandDiff} and \textit{TrajDiff} to $150$ and $300$, respectively.
 \begin{table}[t] 
 
\centering 
\begin{tabular} {@{\hspace{32pt}} c@{\hspace{26pt}}c@{\hspace{26pt}}c@{\hspace{32pt}}  }\toprule
       &  Ours & Interpolation \\ \midrule
     acc. dist. $\downarrow$& \bf{0.379}  & 0.447   \\ \bottomrule
\end{tabular}\caption{\label{tab:accdist_rationet} Wasserstein distances between the acceleration distributions (``acc.~dist'') of ground-truth trajectory and  the generated from \textit{RatioNet} (Ours). We also show the same metric computed on the interpolated subdivided trajectory with an equal length.} 
\end{table}
 {
\renewcommand{\arraystretch}{1.47}
\begin{table}[t]   
\centering \scalebox{0.82}{
\begin{tabular} { ccccc  }\toprule
       &  GT &  Ours &  VAE  & VAEGAN\\   \midrule
 reality score $\uparrow$& 7.10$\pm$2.09 & \bf{6.01}$\pm$2.08 & 5.10$\pm$2.24 &  4.54$\pm$2.39 \\ \bottomrule 
\end{tabular} }\caption{\label{tab:user_study} Results of the user study (perceptual motion quality).} 
\end{table} }
 \begin{figure*}[t!] 
\includegraphics[ width=1\linewidth ]{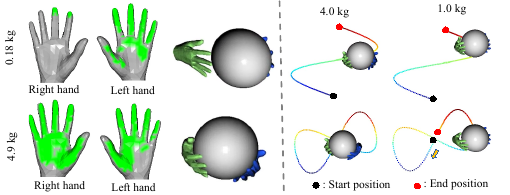} 
\caption{\textbf{(left)} Example visualizations of the contacts synthesized by \textit{ConNet}, given conditioning mass values of 0.18 kg (top) and 4.9 kg (bottom). With heavier mass, the contact region spans the entire palm region whereas contacts concentrate around the fingertips for a light object. \textbf{(right)} Example visualizations of 3D object manipulation given user input trajectories of S curve (top) and infinity curve (bottom). Thanks to the \textit{RatioNet}, the object manipulation speed matches our intuition \ie slower manipulation speed with heavier objects, and vice versa. See our supplementary video for the sequential visualizations.
} \label{fig:contact_userinput} 
\end{figure*}
\section{Further Evaluations}\label{sec:further_evaluations}
In this section, we show further ablative studies to evaluate the significance of the components in our method.\\
\textbf{Temporal loss terms $\mathcal{L}_{vel.}$ and $\mathcal{L}_{acc.}$}: 
 to report the ablative study of the loss terms $\mathcal{L}_{vel.}$ and $\mathcal{L}_{acc.}$ for the network training, we compute the Wasserstein distance between the accelerations of the sampled data and the \GT{} data denoted as ``acc. dist.'' in Table \ref{tab:accdist}. Combining the two loss terms $\mathcal{L}_{vel.}$ and $\mathcal{L}_{acc.}$, our method shows the shortest distance from the \GT{} acceleration distributions.\\
\textbf{Plausibility of the conditioning mass value effect}: can be evaluated by measuring the similarity between the GT object accelerations and the sampled ones. In Table \ref{tab:accdist_conditioning}, we show the ``acc. dist.'' between the accelerations of the ground truth object motions and the sampled motions \textit{with and without} mass conditioning. With the conditioning mass value, our network synthesizes the motions with more physically plausible accelerations on each mass value compared with the network without mass conditioning. \\
\textbf{Effect of RatioNet on the user-provided trajectories}: The goal of RatioNet is to provide plausible dynamics on the user-provided trajectories given conditioning mass values \eg higher object motion speed appears with lighter mass and the object is moved slower with heavier mass value. For the ablative study of RatioNet, we report the ``acc. dist.'' with and without RatioNet comparing with the acceleration distributions of our GT trajectories. For the component without RatioNet, we simply apply uniform sampling on the provided trajectories, denoted as ``Interpolation'' in Table \ref{tab:accdist_rationet}. Thanks to our RatioNet, the object acceleration shows much more plausible values than without the network, faithfully responding to the conditioning mass values. The qualitative results of RatioNet can be seen in our supplementary video.\\
\begin{figure}[t!] 
\includegraphics[ width=1\linewidth ]{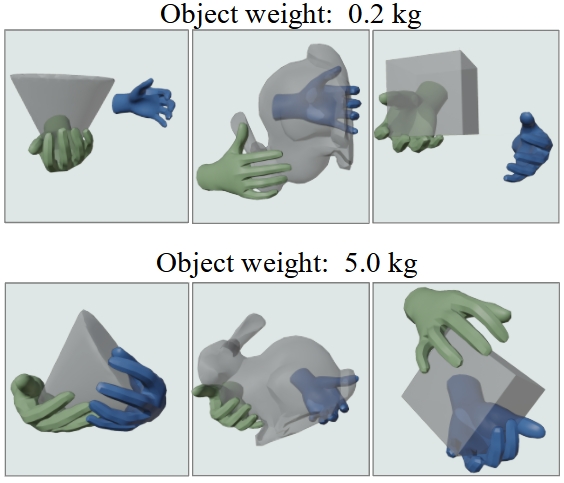} 
\caption{  
Example visualizations of 3D manipulations of the objects unseen during the training, given conditioning mass value of $0.2$kg (top) and $5.0$kg (bottom). \method\ adapts to unseen shapes thanks to its mass-conditioned synthesized hand contacts. 
} \label{fig:dif_shapes} 
\end{figure}
\paragraph{User Study} The realism of 3D motions can be perceived differently depending on individuals. To quantitatively measure the plausibility of the synthesized motions, we perform an online user study. We prepared $26$ questions with videos and gathered $42$ participants in total. The videos for the study were randomly selected from the sampled results of VAE and VAEGAN baselines, \method\ and the \GT{} motions. In the first section, the subjects were asked to select the naturalness of the motions on a scale of $1$ to $10$ \textit{reality score} ($1$ for completely unnatural and $10$ for very natural). Table \ref{tab:user_study} shows the mean scores. \method\ clearly outperforms other benchmarks in this perceptual user study, thanks to our diffusion-based networks that synthesize 3D manipulations with high-frequency details. In the additional section, we further evaluated our method regarding how faithfully the synthesized motions are affected by the conditional mass value. We show two videos of motions at once where the network is conditioned by mass values of $1.0$ and $5.0$, respectively. The participants were instructed to determine which sequence appeared to depict the manipulation of a heavier object. On average, the participants selected the correct answer with $92.8 \%$ accuracy, which suggests that \method\ plausibly reflects the conditioning mass value in its motion. 
 
\paragraph{Qualitative Results}
 In Fig.~\ref{fig:contact_userinput} - (left), we provide visual examples of synthesized contacts with different mass values (0.18kg and 4.9kg). The synthesized contacts are distributed across the palm region when a heavier mass is given, whereas they concentrate around the fingertips with a lighter mass, which follows our intuition. Additionally, Fig.~\ref{fig:contact_userinput} - (right) displays example synthesis results with user-provided input trajectories (S-curve and infinity curve). Thanks to the RatioNet, the object speed reflects the conditioning mass value, \ie faster speed for lighter mass and vice versa. See our supplementary video for its sequential visualizations.

\paragraph{Unseen Objects}
In Fig.~\ref{fig:dif_shapes}, we show the synthesized motions for objects that were not seen during the training, specifically a cone, the Stanford bunny and a cube. Thanks to the synthesized hand contact labels conditioned by a mass value, \method\ shows modest adaptations to different shapes while still correctly reflecting the provided mass values. 